\documentclass{article}

 \usepackage[preprint]{neurips_2025}

 \usepackage{xcolor}
\usepackage{listings}
\usepackage{amsmath}

\lstdefinestyle{prompt}{
  basicstyle=\ttfamily\small,
  backgroundcolor=\color{black!5},
  frame=single,
  breaklines=true,
  columns=fullflexible
}

\usepackage[utf8]{inputenc} 
\usepackage[T1]{fontenc}    
\usepackage{hyperref}       
\usepackage{url}            
\usepackage{booktabs}       
\usepackage{amsfonts}       
\usepackage{nicefrac}       
\usepackage{microtype}      
\usepackage{xcolor}         
\usepackage{array} 
\usepackage{multirow}
\usepackage{graphicx}
\usepackage[]{todonotes}
\usepackage{wrapfig} 
\usepackage{tabularx} 
\usepackage{makecell} 

\title{CurLL: A Developmental Framework to Evaluate Continual Learning in Language Models}

\author{%
  Pavan Kalyan \\
  Microsoft Research \\
  \texttt{tankalapavankalyan@gmail.com} \\
  \And
  Shubhra Mishra \\
  KTH Royal Institute of Technology \\
  Stockholm, Sweden \\
  \texttt{shubhram@kth.se} \\
  \And
  Satya Lokam \\
  Microsoft Research \\
  \texttt{satya.lokam@microsoft.com} \\
  \And
  Navin Goyal \\
  Microsoft Research \\
  \texttt{navingo@microsoft.com} \\
}

\newcommand{\dataset}{\textsc{CurLL}}
\begin{document}

\maketitle

\begin{abstract}
We introduce a comprehensive continual learning dataset and benchmark (\dataset) grounded in human developmental trajectories from ages 5-10, enabling systematic and fine-grained assessment of models' ability to progressively acquire new skills. {\dataset} spans five developmental stages (0-4) covering ages 5-10, supported by a skill graph that breaks down broad skills into smaller abilities, concrete goals, and measurable indicators, while also capturing which abilities build on others. We generate a 23.4B-token synthetic dataset with controlled skill progression, vocabulary complexity, and format diversity, comprising paragraphs, comprehension-based QA (CQA), skill-testing QA (CSQA), and instruction-response (IR) pairs. Stage-wise token counts range from 2.12B to 6.78B tokens, supporting precise analysis of forgetting, forward transfer, and backward transfer. Using a 135M-parameter transformer trained under independent, joint, and sequential (continual) setups, we show trade-offs in skill retention and transfer efficiency. By mirroring human learning patterns and providing fine-grained control over skill dependencies, this work advances continual learning evaluations for language models.
\end{abstract}

\section{Introduction}

The ability to continuously learn and adapt to new information throughout life is one of the hallmarks of human intelligence. Unlike current artificial intelligence systems (e.g., LLMs, agents), humans integrate new knowledge with existing understanding, build increasingly complex skills on earlier foundations, and retain previous capabilities even as they master new ones, and achieve all this with very high sample efficiency. This capacity for lifelong learning represents not just a practical advantage but a fundamental aspect of intelligence itself~\citep{Kudithipudi2022BiologicalUFA, Yan2024PromisesACA, Schmidgall2023BraininspiredLIA}. The continual learning (CL) problem thus is one of the grand challenges for achieving human-like artificial intelligence. It addresses the core problem of how computational systems can progressively acquire, integrate, and refine knowledge over extended periods without compromising earlier capabilities. For  language models (LMs), this challenge is particularly interesting: despite their impressive performance across various tasks, these models face a fundamental limitation in that their skill-set and knowledge of the world becomes static after training, frozen at the point of deployment~\citep{Shi2024ContinualLOA, Wu2024ContinualLFA, Bell2025TheFOA}. In real world, this information continually expands and updates, and this limitation poses a significant challenge to the long-term utility and relevance of LMs.

Despite the importance of the continual learning problem for LMs, current evaluation methodologies suffer from significant limitations:
\begin{enumerate}
    \item Poor skill control: Existing benchmarks often lack precise control over the specific skills being tested, making it difficult to isolate the effects of learning new capabilities~\citep{Liu2025EvaluatingTPA, Rivera2022LatentPOA}.
    \item Unclear knowledge dependencies: The relationships between skills are rarely explicitly modeled, missing out on important transfer effects~\citep{Zheng2025LifelongAgentBenchELA, Nekoei2021ContinuousCAA}.
    \item Inadequate forgetting metrics: Many evaluations fail to properly measure catastrophic forgetting across sequential learning tasks~\citep{Chen2023IsFLA, Huang2023ACSA}.
\end{enumerate}

These limitations make it difficult to understand to measure the efficacy of continual learning algorithms for LMs. This in turn impedes the development of more effective algorithms. 
\begin{figure*}[t] 
    \centering
    \includegraphics[width=1\linewidth]{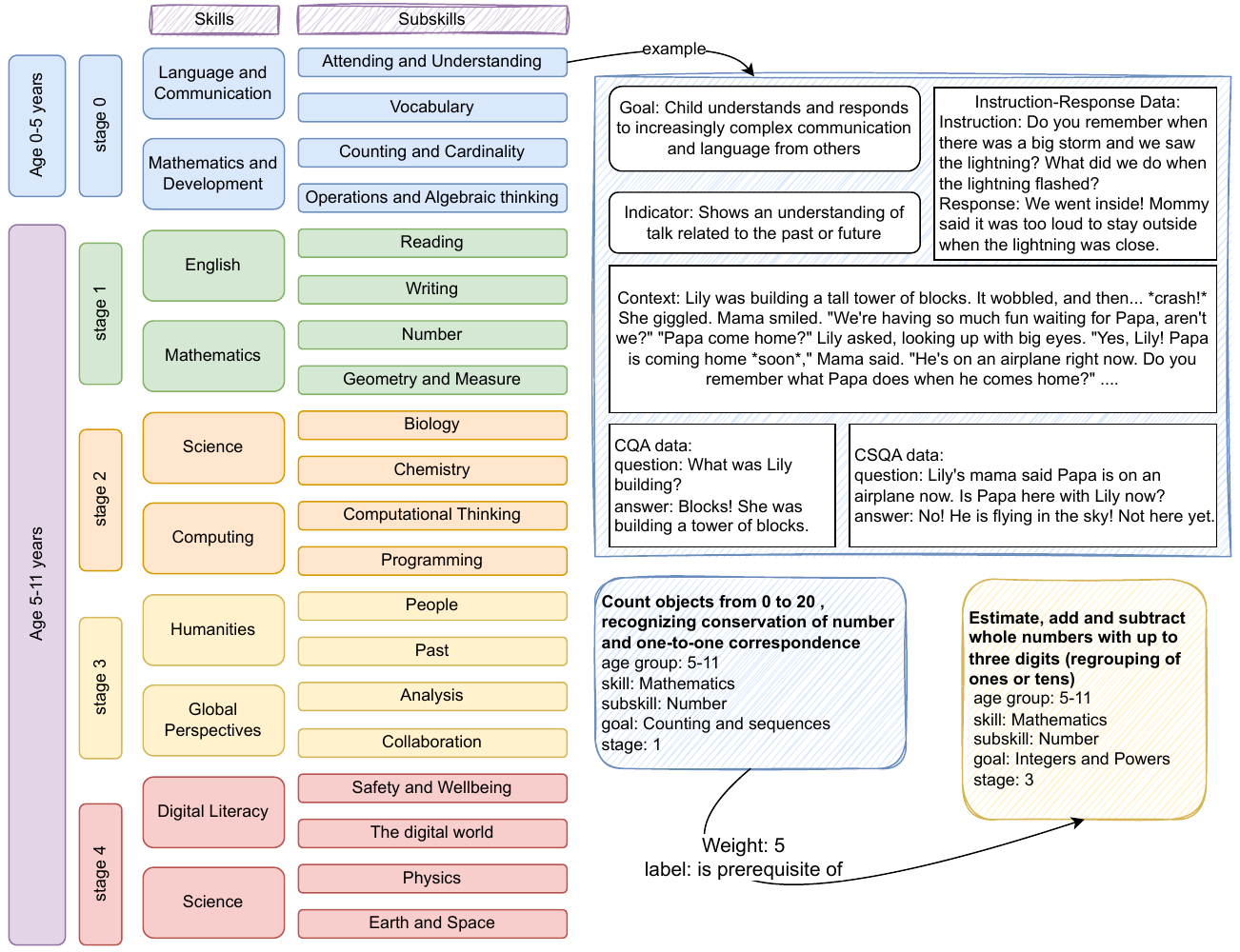}
    \vspace{-2mm} 
    \caption{Developmental framework for children aged 0-11 years, categorized into stages (0-4). Only examples of skills and subskills are mentioned here. An example of how the data looks like is given in the top right. Two nodes and an edge from the skill graph is given in the bottom right.}
    \label{fig:placeholder}
    \vspace{-2mm} 
\end{figure*}

To address these gaps, we introduce a dataset (\dataset) to train and evaluate continual learning algorithms for language models. Coming up with a set of skills with a rich structure and dependencies is a challenge in the construction of such a dataset. We find such a source of skills in human education. \dataset{} is grounded in the curriculum for human education from ages 5-10, divided into five developmental stages (0-4). Each of these stages represent one human-year. Our framework incorporates 1,300+ fine-grained skills. The dependencies among these skills are codified in a skill graph having skills as nodes with the edges capturing a prerequisite relationship. The edges are weighted on a scale of (1-5) to capture dependency strength. Starting from this set of skills, we generate a synthetic dataset of 23.4B tokens, with controlled vocabulary complexity (stage-specific word sampling from Age-of-Acquisition data as seed) and multiple formats (paragraphs, comprehension QA, skill-testing QA, instruction-response). Each stage's dataset ranges from 2.12B to 6.78B tokens, enabling fine-grained evaluation at indicator, skill, and stage levels. Our code\footnote{Training code and trained models: https://github.com/tpavankalyan/CurLL-training}, dataset (stages 0-4), and skill graph\footnote{Data pipeline and generated data: https://github.com/tpavankalyan/CurLL-DataPipeline} is publicly released. Our contributions include:
\begin{itemize}
    \item The idea of grounding skills in human education curriculum in the context of CL.
    \item A synthetic data generation pipeline spanning 5 developmental stages with stage-specific vocabulary and explicit skill dependencies. This pipeline gives us a benchmark with fine-grained control over measuring skill transfer, forgetting and sample efficiency 
    \item A skill graph-based dependency model that explicitly captures prerequisite relationships between learning objectives, enabling nuanced analysis of skill transfer and forgetting
\end{itemize}

\section{Related Work}
One particular limitation of LMs is that their knowledge is confined to fixed parameters established during training~\citep{Du2023FromST}. While LLMs encode world knowledge in their parameters through pretraining, this knowledge can quickly become outdated as the world changes~\citep{Jang2021TowardsCK}. Continual learning techniques address this by enabling models to learn continually and adapt, integrating new knowledge and skills while retaining previously learned information~\citep{Zheng2024TowardsLL}. This capability represents a fundamental property of human intelligence: the capacity to dynamically adapt cognition by ingesting new knowledge from the environment over time~\citep{Du2023FromST, Jin2021LifelongPC}. The core challenge in continual learning is the stability-plasticity dilemma, which requires models to balance the previous skills (stability) with the ability to learn new tasks (plasticity)~\citep{Jiang2024RefineLL, Wang2025LifelongLW, Liu2025LowComplexityII}. Catastrophic forgetting emerges as the primary manifestation of this challenge, where LMs tend to forget previously acquired knowledge when learning new instances~\citep{Huang2024MitigatingCF, Zeng2023ContinualLW, Liao2025DATADA}.

Many datasets and benchmarks exist for continual learning of language models such as TRACE~\citep{Wang2023TRACEAC}, MMLM-CL~\citep{Zhao2025MLLMCLCLA}, OCKL~\citep{Wu2023OnlineCKA}, CKL~\citep{Jang2021TowardsCKA}, TemporalWiki~\citep{Jang2022TemporalWikiALA} and TiC-LM~\citep{Li2025TiCLMAWA} etc. TRACE~\citep{Wang2023TRACEAC} highlights the problem in existing benchmarks that are often too simple or are already included in the LLM instruction-tuning sets. It also introduces new metrics to evaluate shift in LLM abilities. MMLM-CL~\citep{Zhao2025MLLMCLCLA} discusses the shortcomings of current Cl benchmarks as lack of real world applicability and IID evaluation. OCKL~\citep{Wu2023OnlineCKA} proposes new metrics for measuring knowledge acquisition rate and knowledge gap but concentrates primarily on knowledge-intensive tasks as compared to procedural tasks.  TemporalWiki~\citep{Jang2022TemporalWikiALA} also concentrates on updating factual information in language models based on temporal data constructed from Wikipedia snapshots. Several domain-specific benchmarks exist as well for language models. Continual relation extraction has been evaluated on datasets including Continual-FewRel, Continual-SimpleQuestions, and Continual-TACRED, where relations are partitioned into sequential tasks~\citep{Wu2021CurriculumMetaLF}. SuperNI contains a variety of traditional NLP tasks and serves as a practical benchmark for continual learning of large language models~\citep{He2024SEEKRSA}. \citet{Yang2024MoRALMA} introduced the Life Long Learning of LLM (5L-bench) benchmark, which encompasses a curated dataset of question-answer pairs and evaluation metrics for both open-book and closed-book settings.

Despite these developments, existing continual learning benchmarks are often considered unsuitable for evaluating state-of-the-art LMs~\citep{Wang2023TRACEAC, Razdaibiedina2023ProgressivePC, Scialom2022FinetunedLM, Zhang2015CharacterlevelCN}. These benchmarks often emphasize artificial task boundaries~\cite{He2024SEEKRSA}, lack temporal and distributional complexity. Moreover, these datasets do not offer precise control over skills or information to validate the effectiveness of existing solutions for continual learning. Skill-it~\citep{Chen2023SkillitADA} introduces a data sampling algorithm for continual pretraining and finetuning. They do this by arranging the skills in a increasing order of complexity. Other existing works~\citep{Khetarpal2020TowardsCRA, Greco2019PsycholinguisticsMCA, Xu2024KlFKLA} discuss the importance of skill distinction and its effect on evaluating continual learning. 

In contrast, our benchmark is grounded in human developmental curricula and provides a large-scale synthetic dataset with explicit skill dependencies and controlled progression, enabling fine-grained evaluation of transfer, forgetting, and sample efficiency beyond what existing benchmarks support.

\section{Dataset Setup}
One of the main design decisions in our dataset construction is to precisely specify the skills that the model learns at each stage of CL. To this end, our framework for evaluating continual learning is grounded in human learning curriculum, with the dataset designed to mimic the developmental stages from age 5-10. This section details our methodology for constructing the dataset, developing the skill graph that models dependencies between skills, and creating test-train splits for evaluation.
\subsection{Grounding in Human Curricula}
We use two established educational frameworks to develop our skill taxonomy: the Early learning Outcomes framework (ELOF)\footnote{\citet{ELOF_2024}} for children below age 5 and the Cambridge curriculum\footnote{\citet{Cambridge_2025}} for children aged 5-10. These frameworks help us define fine grained notion of skills as specified by a skill-tuple that consists of four components:
\begin{itemize}
    \item Skills\footnote{"Skill" here has a specific meaning, which is different from the general notion of skill used before}: High-level domains or subjects (e.g. Mathematics, Science)
    \item Sub-skills: Specific components within a skill (e.g., Counting and Cardinality)
    \item Goals: Broad statement of learning expectations within a sub-skill
    \item Indicators: Specific, observable behaviors that demonstrate mastery of a goal
\end{itemize}
Examples of each of these can be seen in Figure~\ref{fig:placeholder}. 

The ELOF framework, introduced by the U.S. Office of Head Start in 2015, provides a comprehensive roadmap for child development from birth to age five across five broad areas: Approaches to Learning, Social and Emotional Development, Language and Literacy, Cognition, and Perceptual, Motor, and Physical Development. For ages 5-11, we use the Cambridge Primary Curriculum, which covers subjects including English, Mathematics, Science, Computing, and Global Perspectives. The curriculum structure flows from subjects (renamed as skills in our framework) to domains/strands (renamed as subskills), then to substrands (goals), each with specific learning objectives (indicators). We also adopt the notion of stages from the Cambridge curriculum in our framework, where each stage corresponds to one year starting from age 5. Therefore, we have 5 stages in our framework, where stage 0 denotes ages up to 5, stage 1 denotes age 5-6 and so on. The number of skill-tuples in our framework is same as the number of indicators present up to stage 4, statistics of which are mentioned in Table~\ref{tab:stage-stats}.
\begin{wrapfigure}{r}{0.5\linewidth}  
    \centering
    \includegraphics[width=1\linewidth]{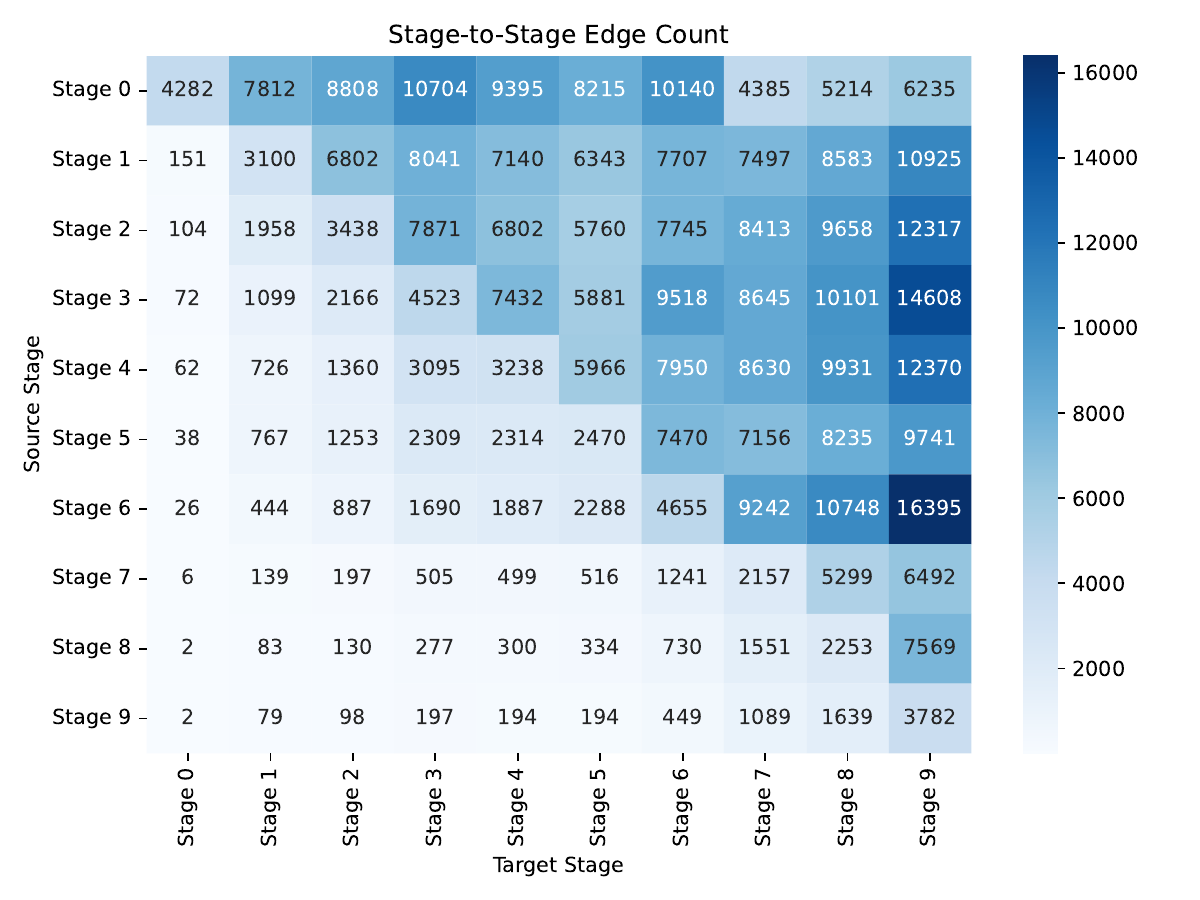}
    \caption{Heatmap showing the number of prerequisite edges in the skill graph. Rows correspond to source stages, columns to target stages, and color intensity indicates the number of connections.}
    \label{fig:skill_graph}
    \vspace{-7mm} 
\end{wrapfigure}

\subsection{Skill Graph}
A critical component of our framework is the skill graph, which captures the prerequisite relationships between indicators. This is a directed graph that has indicators as nodes, with edges representing prerequisite relationships weighted from 1-5 to indicate dependency strength. These relationships model how skills are built on each other in developmental stages. We use an LLM\footnote{Gemma3-27B-IT is used for all LLM inferences throughout this work} to predict these dependency relationships between indicators. While the skill graph isn't directly used for skill data generation, it provides insights for analyzing continual learning patterns and interpreting evaluation results. To validate the skill graph, we examine the distribution of incoming and outgoing edges across different stages. As expected, lower stages generally exhibit fewer incoming dependencies, while higher stages accumulate more prerequisite relationships. Figure~\ref{fig:skill_graph} illustrates this trend: the number of edges flowing from lower to higher stages is substantially greater than the reverse. This aligns with the progressive nature of the stages, where foundational skills at earlier stages support the development of more advanced skills later on. A small fraction of edges, however, point from higher to lower stages. These arise from limitations in our edge prediction method, which relies on LLM-based inference over skill tuples. For instance, one such erroneous edge connects an English skill in stage 2 to a Science skill in stage 1.

\begin{wrapfigure}{r}{0.5\linewidth}  
    \centering
    \includegraphics[width=1\linewidth]{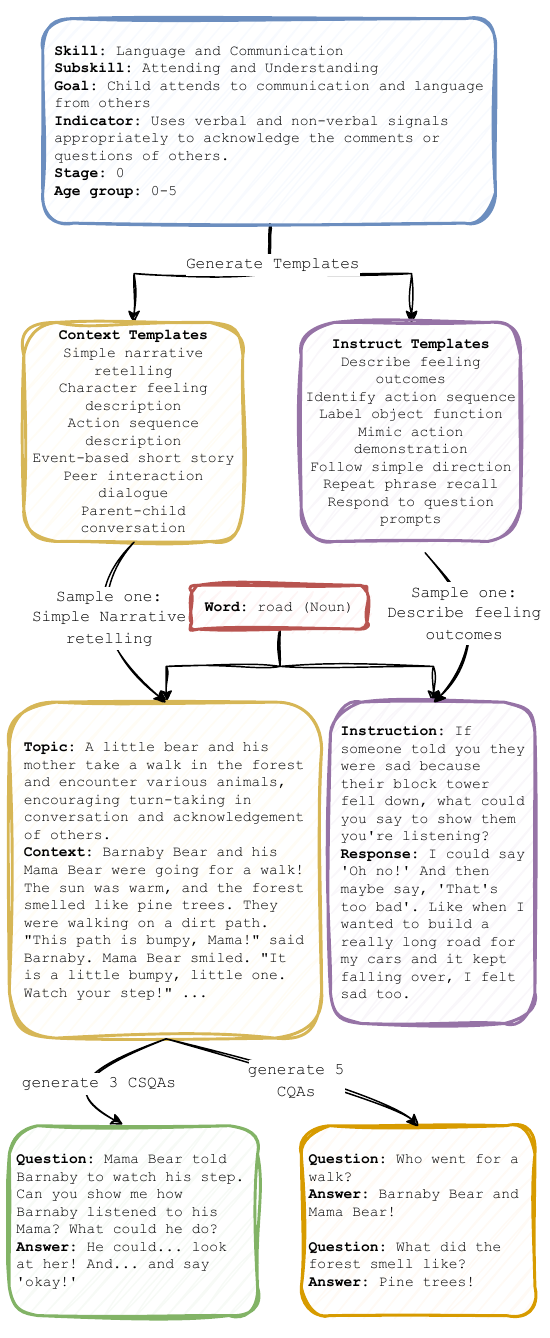}
    \caption{Synthetic data generation pipeline}
    \label{fig:llm flow}
    \vspace{-10mm}
\end{wrapfigure}
\subsection{Synthetic Data Generation}
Our synthetic data consists of instances, each mimicking a situation a child might encounter. Instances are of three types: (1) IR: an instruction-response pair, where the instruction is about some general world knowledge, (2) CQA: context-based question-answers for testing comprehension, (3) CSQA: context-based question-answers for testing skills. A context is a short piece of text which forms the basis of the corresponding question-answer pairs in the instance. Contexts can be of multiple types as specified by a template: for example, a simple narrative, or a dialogue. Similarly, IR-pairs can also have different types specified by templates, e.g.,  mimic action or follow simple direction. See Figure~\ref{fig:llm flow} for more examples.

Instances are generated by prompting an LLM with a \textit{seed}. A seed consists of a skill-tuple, vocabulary seed, instance type, template. This choice is crucial for ensuring diversity and coverage of our data.
An example of our seed and the generated instance is given in Figure~\ref{fig:llm flow}.
This tuple is also our way to ground the generations in the skill graph. 
In more detail, a seed is generated as follows:
\begin{enumerate}
    \item Age-appropriate skill grounding: Each generated instance is tied to a specific skill-tuple from our curriculum framework. Since this tuple contains the stage and age group, the generated data is expected to be grounded in the same.
    \item Vocabulary seed: To generate the data at scale, we use additional seeds for diversity. One of them is the words from vocabulary of a child belonging to a stage in the curriculum. We do this by using the Age-of-Acquisition data~\citep{Kuperman2012AoA}, where words along with the age-rating based on human studies are presented. 1000 words are sampled for each stage. A vocabulary seed consists of one randomly chosen word from this list. 
    \item One of the types (IR, CQA, CSQA). 
    \item Templates: For each skill-tuple, we generate at least 15 templates for contexts and for IR-pairs. Therefore, each skill tuple has at least 15 types of context templates like stories, dialogue, etc. and 15 types of instruction-response templates like why questions, describing the event, etc. Examples of these templates are mentioned in Figure~\ref{fig:llm flow}. We use an LLM to generate these templates by giving the skill-tuple as the input. The prompts for generating context and IR templates are given in Appendix (\ref{app: templates}).
\end{enumerate}

To generate one instance of the data, we first construct a seed: each skill-tuple is combined with a vocabulary seed for that stage, an instance type, and a template for that instance type. If the instance type is CQA or CSQA\footnote{While CQA, CSQA are very similar to each other, CQA is built to test the language understanding abilities of the model (which in itself is a skill). This acts as a better replacement for perplexity to measure the performance of a chat like model.}, then we first generate the context and then using the context, we generate the corresponding question-answers. If the instance type is IR, then we directly generate the instruction-response pairs. The prompts for all generations are presented in Appendix~\ref{app:prompts}. 
In our dataset, each instance includes the seed used to generate it as part of its metadata.
\begin{figure}[h]
    \centering
    \includegraphics[width=1\linewidth]{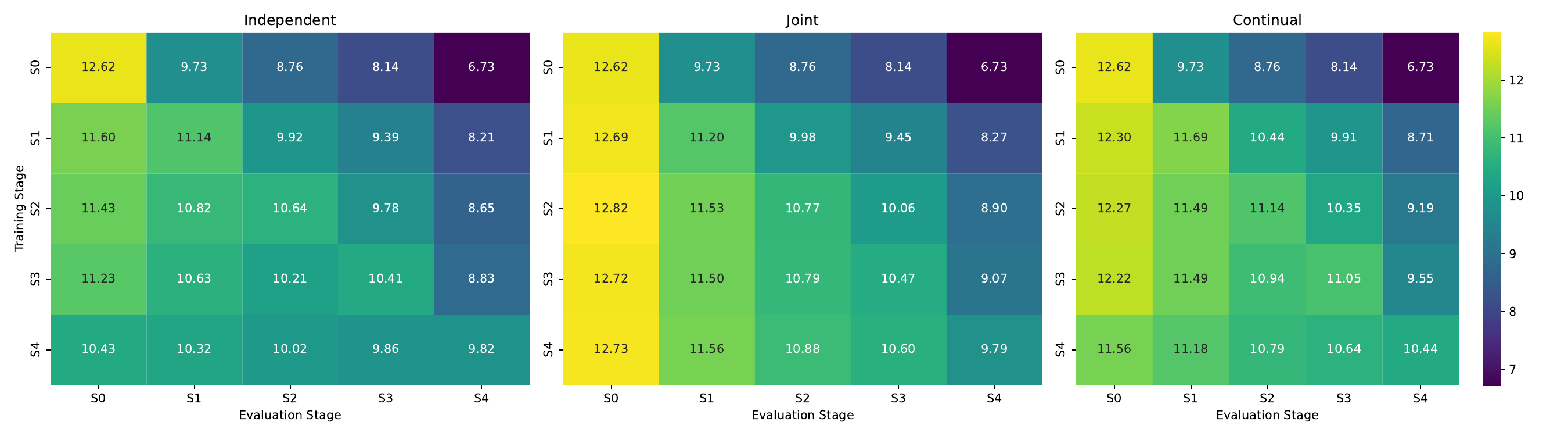}
    \caption{Stage-wise evaluation results for different training setups. Independent training corresponds to models trained on a single stage, joint training to models trained on mixtures of data up to a stage, and continual training to sequential upto a stage. Heatmaps report summed correctness scores across all test formats (IR, CQA, CSQA)}
    \label{fig:main result}
\end{figure}
\begin{table}[t]
\small
\centering
\setlength{\tabcolsep}{4pt}
\renewcommand{\arraystretch}{1.1}
\caption{Dataset statistics across developmental stages (0-4), including total tokens}
\begin{tabularx}{\linewidth}{c|cccc|ccc|X} 
\hline
\multirow{2}{*}{\textbf{Stage}} 
& \multicolumn{4}{c|}{\textbf{Skills \& Goals}} 
& \multicolumn{3}{c|}{\textbf{Instances}} 
& \multirow{2}{*}{\makecell{\textbf{\# Tokens}\\\textbf{in Bn}}} \\
\cline{2-8}
& \textbf{\# Skills} & \textbf{\# Sub-skills} & \textbf{\# Goals} & \textbf{\# Indicators} 
& \textbf{\# CQA} & \textbf{\# CSQA} & \textbf{\# IR Pairs} &  \\
\hline
0 & 7  & 24  & 59  & 182  & 1.0M  & 3.01M  & 3.30M  &  2.12 \\
1 & 7  & 29  & 86  & 292  & 20.2M  & 4.04M  & 4.10M  &  3.47 \\
2 & 6  & 26  & 67  & 249  & 23.5M  & 4.70M  & 4.78M  &  4.56 \\
3 & 6  & 26  & 68  & 271  & 31.2M  & 6.24M  & 6.29M  &  6.47 \\
4 & 6  & 23  & 70  & 349  & 27.4M  & 5.49M  & 5.52M  &  6.78 \\
\hline
\end{tabularx}
\label{tab:stage-stats}
\end{table}
\subsection{Data Statistics and Verification} 
\label{sec:verif}
\begin{wraptable}{r}{0.52\textwidth}
\centering
\small
\caption{Diversity and Deduplication metrics}
\vspace{0.5em}
\begin{tabular}{c|cc|cc}
\toprule
\textbf{Stage} & \multicolumn{2}{c|}{\textbf{Context}} & \multicolumn{2}{c}{\textbf{IR}} \\
               & \textbf{Div $\uparrow$} & \textbf{Dedup $\downarrow$} & \textbf{Div $\uparrow$} & \textbf{Dedup $\downarrow$} \\
\midrule
0 & 34.29\% & 11.83\% & 30.77\% & 3.50\% \\
1 & 35.60\% & 5.36\%  & 31.73\% & 3.85\% \\
2 & 34.17\% & 15.47\% & 32.64\% & 2.54\% \\
3 & 34.68\% & 14.86\% & 32.97\% & 2.09\% \\
4 & 35.45\% & 13.41\% & 33.14\% & 1.93\% \\
\bottomrule
\end{tabular}
\label{tab:div-dedup}
\end{wraptable}
We generated data for stages 0 through 4 inclusive, containing a total of 23.4B tokens (Table~\ref{tab:stage-stats}). We use two methods to measure this diversity of generated data: 1) Diversity as reciprocal of compression ratio using gzip \cite{gailly_adler_1992_gzip}. 2) The intra- and inter-text deduplication rate as calculated by semantic deduplication. Details of how these measures are implemented are given in the appendix~\ref{app:verify}. Cross-stage analysis shows higher diversity and lower deduplication rate (<5\%) between stages compared to intra-stage results, confirming that content evolves meaningfully across developmental progression while maintaining stage-specific uniqueness. The results for these methods are presented in Table~\ref{tab:div-dedup}. 

Another important feature of the dataset is the progression in the difficulty of the skills as the stage number increases. We sample 500K instances from each stage for each data type and run statistical readability tests\footnote{These tests use pre-defined word corpuses to predict the grade a text belongs to. We use the following repo to measure the readability: https://github.com/cdimascio/py-readability-metrics}. Means across multiple readability metrics are reported in Table~\ref{tab:readability}. The readability tests show that as stages progress, the texts also become increasingly challenging. At least 50 random instances from each dataset per stage were manually analysed. Based on manual inspection, CQA data for all stages was found to be accurate. IR and CSQA data had certain patterns like excessive use of discourse markers for early stages and verbose response to instructions. 

\begin{wraptable}{l}{0.52\textwidth}
\centering
\begin{tabular}{l|cccc}
\toprule
\textbf{Stage} & \textbf{Context} & \textbf{CQA} & \textbf{CSQA} & \textbf{IR} \\
\midrule
0 & 4.61 \tiny{1.87} & 2.38 \tiny{2.88} & 3.07 \tiny{2.26} & 4.48 \tiny{1.52} \\
1 & 5.24 \tiny{1.72} & 4.39 \tiny{1.81} & 4.44 \tiny{1.62} & 4.86 \tiny{1.41} \\
2 & 5.18 \tiny{1.93} & 4.39 \tiny{1.80} & 4.69 \tiny{1.54} & 4.69 \tiny{1.59} \\
3 & 5.51 \tiny{1.85} & 4.65 \tiny{1.70} & 4.98 \tiny{1.46} & 5.03 \tiny{1.50} \\
4 & 6.42 \tiny{1.79} & 5.63 \tiny{1.44} & 5.96 \tiny{1.30} & 5.91 \tiny{1.34} \\
\bottomrule
\end{tabular}
\caption{Average readability scores across stages.}
\label{tab:readability}
\end{wraptable}

\subsection{Train-test split}
We aim to keep the size of the test set per stage to be between 5k-7k samples. And to ensure uniform coverage of indicators in the test data, we choose 25 instances per indicator. Since the data is synthetically generated at scale, though we cannot validate the entire data, we reserve the highest quality samples for the test set. 
Three types of test sets are created based on the three instance types. 100 random samples are selected per indicator for each of the data format. All 100 questions are graded by the LLM on a scale of 1-5 for correctness of the response. The detailed rubrics and prompts are presented in Appendix~\ref{app:prompts}. 25 highest scoring (mostly 5) instances out of all instructions were selected in the test set. 25 randomly selected instances from the remaining samples were put in the validation split. And all the rest of the instances remain in the training set. Note that this construction ensures that the test, validation and training sets are disjoint because they use distinct seeds. 

\section{Experiments and Results}
\begin{figure}
    \centering
    \includegraphics[width=1\linewidth]{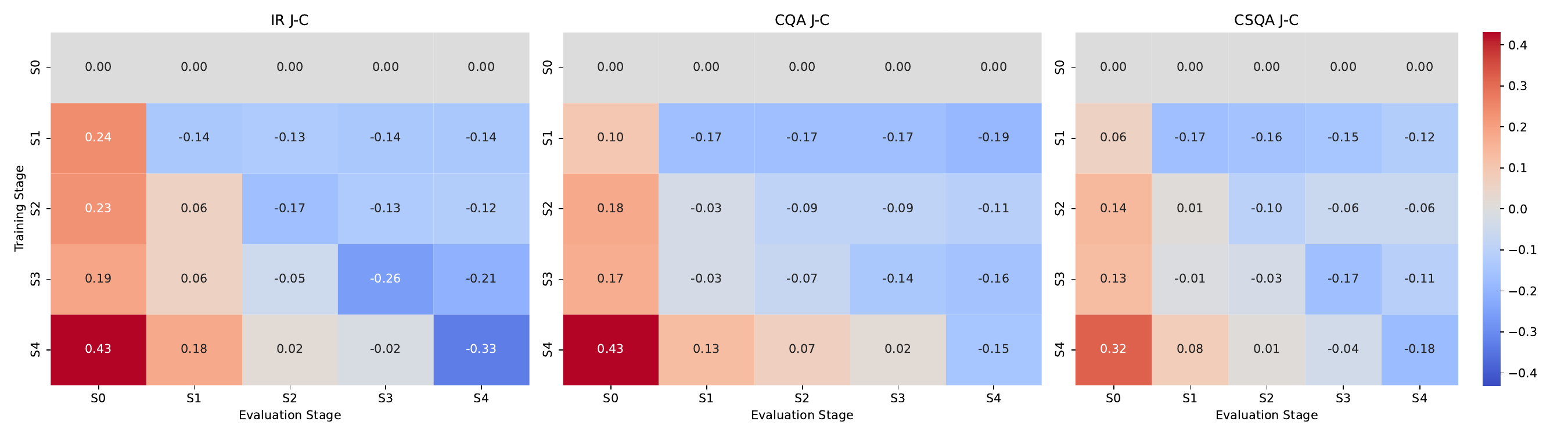}
    \caption{Forgetting analysis across training setups. The plots show performance differences between joint and continual training for IR, CQA, and CSQA test sets across stages 0-4. The Y-axis corresponds to models trained upto a stage. The X-axis corresponds to test set of mentioned stage.}
    \label{fig:j_c}
\end{figure}
We do preliminary experiments to validate the quality of our dataset (in addition to the checks performed in Section~\ref{sec:verif}) that exposes meaningful challenges: whether models can retain earlier-learned skills, how sequential training affects generalization, and to what extent transfer across related skills occurs. By analyzing these at the granularity of skills, we demonstrate that \dataset{} enables insights that are not visible in existing benchmarks. Unlike traditional language model training that includes two stages: pretraining and then finetuning, we do a single phase training. All instance types i.e. CQA, CSQA, and IR are included in the same phase. Since all of them are question-answers, with and without context paragraphs, we use a standard chat template to train the language models from scratch. Smollm2-135M parameter model is used as the base architecture. All training runs are performed on one full epoch of the data. Learning rate of 5e-3 and effective batch size of 1536 instances remain unchanged across experiments. We use a context length of 1024. Other training and inference related hyper-parameters are mentioned in Appendix~\ref{app:hype}. Our preliminary experiments includes three types of training:
\begin{enumerate}
    \item Independent: The model is trained from scratch on data of each stage independently. The models trained in this fashion are denoted by $M_i$ if trained on data of stage $i$.
    \item Joint: Jointly trained on a mixture of stages. The data from different stages is combined and shuffled randomly. $M_{ij}$ denotes a jointly trained model on stage $i$ and stage $j$.
    \item Continual: The model is first trained on stage $i$, then stage $j$, then stage $k$ and so on. $M_{i-j}$ denotes model trained until stage $j$, $M_{i-j-k}$ if it is trained until stage $k$ and so on.
\end{enumerate}

\subsection{Results}
\label{sec:res}
To evaluate the trained models, the instances from test set are passed through the chat template and the model is asked to complete the generation post instruction. These inferences along with the prompt is passed to an LLM to rate on a scale of 1-5. This is followed for all three types of test sets. Each model is evaluated on test sets of all stages. The prompt and rubrics of evaluation are mentioned in Appendix~\ref{app:prompts}. The main objective of the rating is to evaluate the correctness of the model inference with some weightage to the stage on which the model is being evaluated. The summation of scores across test set types (IR, CQA, CSQA) is presented in Figure~\ref{fig:main result}. The individual scores are available in the Appendix. 
\begin{wrapfigure}{r}{0.5\linewidth}  
    \centering
    \includegraphics[width=1\linewidth]{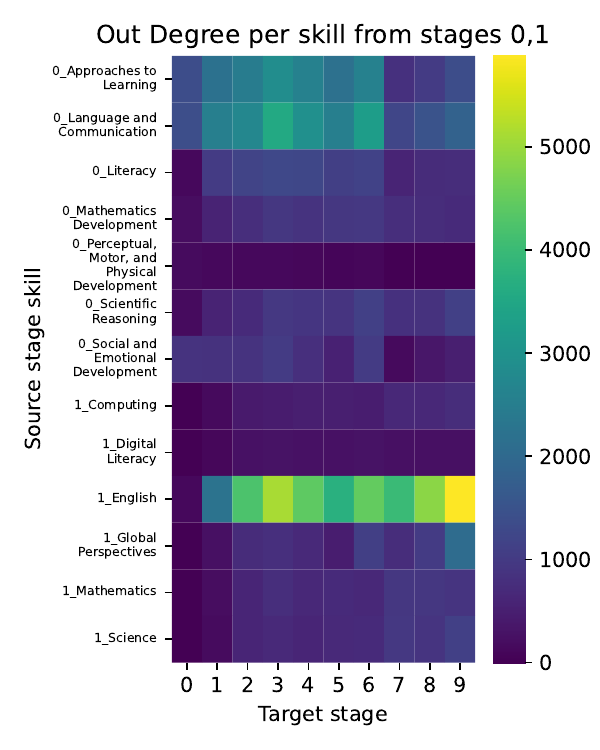}
    \caption{Out-degree distribution of skills from stages 0 and 1 in the skill graph. Skills with fewer outgoing edges (e.g., Perceptual, Motor, and Physical Development; Digital Literacy) are less connected to later stages and are observed to be more vulnerable to forgetting in continual training.}
    \label{fig:out_degree}
    \vspace{-5mm}
\end{wrapfigure}
The Y-axis of the figure shows the stage on which the model is being evaluated. The Y-axis denotes the data used to train the model. For the case of independent training, $S_i$ denotes data from stage $i$ is used for training. For Joint training $S_i$ denotes mixture of data from all stages until $i$ including stage $i$. For Continual training, this means model trained sequentially until stage $i$. 

The figure shows that as compared to independent models ($M_i$), joint models($M_{ij}$) show better generalization to later stages but also, stronger performance on trained stages. However continual models ($M_{i-j}$) shows the best performance on later stages but the performance degrades on already trained stages. This is better shown in Figure~\ref{fig:j_c} across different test set types. Even though $M_{ij}$ and $M_{i-j}$ are trained on exactly on the same amount of data with same hyper-parameter settings, just by changing the order of the data, i.e. by arranging the data in a progressive fashion, leads to better generalization. However this also leads to forgetting of previous skills, which in this case is counter-intuitive as the later skills require mastery of foundational skills.

This is however decoded by referring to the skill-graph. The skill that has the highest difference between the performance of joint and continually trained model for stage 0 ($M_{01}$ vs. $M_{0-1}$) is "Perceptual, Motor, and Physical Development" and for stage 1 ($M_{012}$ vs. $M_{0-1-2}$) is "Digital Literacy". These skills are also the skills that are having the least number of outgoing edges, i.e. all indicators that belong to these skills are rarely prerequisite of future skills\footnote{Perceptual, Motor and Physical Development can be seen as a fundamental skill which one might expect to have more outgoing edges than seen in Figure~\ref{fig:out_degree}. This is explained by the fact that all skills except stage 0 skills are derived from an academic curriculum, while stage 0 refers to skills required for holistic development of a 5 year old.}. This can be seen from Figure~\ref{fig:out_degree}, where the sum of all edges from indicators present in source skills (y-axis) to indicators present in target stages (x-axis) is plotted. All results per indicator per stage are given in the Appendix.

The tasks across stages within the same age group show a high degree of similarity. This is reflected in the strong task transfer observed even when stages are trained independently (\ref{fig:main result}). A similar pattern appears in Table~\ref{tab:readability}, where the average readability scores for stages 1 and 2 in the 5-11 age group are not strictly monotonic. This mirrors how human learning typically progresses: moving from stage 1 to stage 2 usually involves introducing only a few new concepts while increasing the complexity of the concepts already learned. Readability tests, however, capture complexity only in a statistical sense, based on a fixed set of words and sentence structures.

\section{Utility of (\dataset)}
(\dataset) can serve multiple broad purposes to better understand and solve the problem of continual learning of language models. One of the core components is skill graph that can be used as a diagnostic tool. The metadata in {\dataset} allows fine-grained control of the number of instances and skills seen by the model during training. This enables better evaluation of sample efficiency of continual learning algorithms. By leveraging prerequisite edges, one can test whether learning Skill A improves Skill B. As discussed in Section~\ref{sec:res} it also helps interpret forgetting: low-outdegree skills (few dependencies) vs. high-outdegree skills (many dependencies) behave differently. Forgetting, forward transfer, backward transfer, and data efficiency can all be measured at the levels of skills, sub-skills, and indicators. This enables richer analysis than stage- or task-level metrics in existing benchmarks, which typically report only overall accuracy on entire tasks (e.g., classification or QA) without revealing which underlying abilities are gained or lost. The framework also allows data generation at scale, which enables researchers to work on continual pretraining in a much controlled setting as compared to existing works. The work explores stages 0-4 i.e. curriculum only between age 5-9 years. The work can be extended upto age 14 using the skill-graph framework.

\section{Limitations}
A limitation of the present work is that both the instructions and the responses are part of the dataset and the language model ends up learning both. A setup that truly reflects human-like learning would involve, instead of a static dataset, an environment in which the agent learns by interactions. Ultimately, this limitation stems from the nature of language modeling itself rather than being a weakness of data set design. Another limitation of the work is the use of synthetic data exclusively for experiments. While this step was taken to ensure greater control over data, the data might not reflect the real world scenarios of continual learning. Finally, all the experiments are performed on a 135M-parameter model. While perfectly suitable for a proof-of-concept, foundation models are typically orders of magnitude larger. The dynamics of catastrophic forgetting and knowledge transfer may differ significantly at scale. The conclusions drawn from this smaller model may not fully translate to a billion-parameter model.

\section{Conclusion}
We developed a continual learning evaluation framework for language models grounded in human developmental curricula. (\dataset) combines a directed, weighted skill graph of over 1,300 indicators with a 23.4B token synthetic dataset that controls stage-wise vocabulary, difficulty, and format. It enables fine-grained analysis of forgetting at the level of skills, sub-skills, and indicators. Our experiments with independent, joint, and sequential training show that the order of data alone, can affect forgetting and generalization.

\bibliographystyle{plainnat}  
\bibliography{references}


\appendix

\section{Technical Appendices and Supplementary Material}
Technical appendices with additional results, figures, graphs and proofs may be submitted with the paper submission before the full submission deadline (see above), or as a separate PDF in the ZIP file below before the supplementary material deadline. There is no page limit for the technical appendices.


\appendix
\section{Data Verification}
\label{app:verify}
For both the methods, 500K texts are sampled from each of the Paragraphs and Instruction-response pairs.

\subsection{Diversity}
 For the diversity of the text, we follow~\citet{Chang2024ScalingPL} and calculate the compression ratio of the text as 
\[
\mathrm{CR}(D) = \frac{\text{Original size of } D \; (\text{bytes})}{\text{Compressed size of } D \; (\text{bytes})},
\]
and define diversity by
\[
\mathrm{Dr}(D) = 1/\mathrm{CR}(D).
\]
A higher compression ratio $\mathrm{CR}(D)$ indicates greater redundancy, meaning lower diversity in the text.  
Thus, diversity $\mathrm{Dr}(D)$ increases when redundancy decreases. We see diversity ranging between $30.77 \%$ and $35.60\%$, which is similar to other work. Remarkably, despite being constructed to target children at or below age 10, our dataset achieves a diversity level comparable to Wikipedia's English training set $(37.79\%)$. This indicates that the dataset maintains a richness of expression and linguistic variety that is on par with large-scale web text, even though the content is restricted to more cognitively age-appropriate text \cite{wikidump}.

\subsection{Deduplication}
For semantic deduplication\footnote{We use the following repo for semantic deduplication: https://github.com/MinishLab/semhash}, we pass the texts through a sentence encoder and find the deduplication rate as the percentage of sentences that have cosine similarity of at least 0.95 with another sentence in the same stage.

\begin{table}[h]
    \centering
    \caption{Diversity and Deduplication Rates when Considering Pairwise Stages}
    \vspace{0.5em}
    \begin{tabular}{c|cc}

\toprule
\textbf{Stage Pair} & \multicolumn{2}{c}{\textbf{Context}} \\
& \textbf{Div $\uparrow$} & \textbf{Dedup $\downarrow$} \\
\midrule
0, 1 & 31.29\% & 0.3\% \\
0, 2 & 31.96\% & 0.1\%\\
0, 3 & 32.25\% & 0.0\% \\
0, 4 & 32.50\% & 0.0\%\\
1, 2 & 32.27\% & 0.3\%\\
1, 3 & 32.52\% & 0.2\% \\
1, 4 & 32.71\% & 0.1\%\\
2, 3 & 32.82\% & 0.4\% \\
2, 4 & 32.94\% & 0.2\%\\
3, 4 & 33.07\% & 0.2\%\\
\bottomrule
\end{tabular}
    \label{tab:inter-stage-div-dedup}
\end{table}

\subsection{Detailed Readability Metrics}
\label{app:readability}
Note that average grade of the data is slightly higher than the intended age of the data (especially for the first few stages). However, this is because not all skills we generate data for are, in real-life, text-based. Thus, demonstrating them in language ends up requiring complex words, which affects the readability score. For example, children can verbally reason about cause-and-effect in multi-turn conversations, but when written down, that same dialogue is rated at a much higher reading level than the child can actually read, leading to higher readability scores in our data. 
\begin{table}
\small
    \centering
    \setlength{\tabcolsep}{4pt}
\renewcommand{\arraystretch}{1.1}
\caption{Detailed Readability Metrics Across all 5 Stages and Datasets}
\vspace{0.5em} 
    \begin{tabular}{l l c c c c c c}
\toprule
Dataset & Stage & Flesch Kincaid & SMOG & Coleman Liau & Automated Readability & Dale Chall & Gunning Fog \\
\midrule
Context & 0 & 3.15 \tiny{0.35} & 6.90 \tiny{0.76} & 4.28 \tiny{0.51} & 1.68 \tiny{0.47} & 6.69 \tiny{0.27} & 4.94 \tiny{0.34} \\
Context & 1 & 3.68 \tiny{0.35} & 7.55 \tiny{0.73} & 5.18 \tiny{0.50} & 2.58 \tiny{0.49} & 6.70 \tiny{0.29} & 5.74 \tiny{0.35} \\
Context & 2 & 3.80 \tiny{0.36} & 7.54 \tiny{0.75} & 4.21 \tiny{0.46} & 2.25 \tiny{0.48} & 7.18 \tiny{0.35} & 6.12 \tiny{0.38} \\
Context & 3 & 4.16 \tiny{0.36} & 7.84 \tiny{0.74} & 4.58 \tiny{0.48} & 2.71 \tiny{0.50} & 7.27 \tiny{0.36} & 6.48 \tiny{0.38} \\
Context & 4 & 5.13 \tiny{0.42} & 8.76 \tiny{0.79} & 5.39 \tiny{0.51} & 3.77 \tiny{0.56} & 7.89 \tiny{0.34} & 7.58 \tiny{0.45} \\
\midrule
CQA & 0 & 0.79 \tiny{0.35} & 5.06 \tiny{0.59} & 0.26 \tiny{0.59} & -1.47 \tiny{0.43} & 6.89 \tiny{0.30} & 2.75 \tiny{0.34} \\
CQA & 1 & 2.73 \tiny{0.37} & 6.45 \tiny{0.59} & 4.10 \tiny{0.54} & 1.65 \tiny{0.49} & 6.47 \tiny{0.26} & 4.92 \tiny{0.45} \\
CQA & 2 & 2.74 \tiny{0.38} & 6.42 \tiny{0.59} & 4.00 \tiny{0.53} & 1.67 \tiny{0.50} & 6.44 \tiny{0.28} & 5.07 \tiny{0.45} \\
CQA & 3 & 3.04 \tiny{0.37} & 6.66 \tiny{0.59} & 4.37 \tiny{0.52} & 2.08 \tiny{0.49} & 6.41 \tiny{0.27} & 5.36 \tiny{0.44} \\
CQA & 4 & 4.08 \tiny{0.38} & 7.54 \tiny{0.59} & 5.59 \tiny{0.48} & 3.52 \tiny{0.49} & 6.50 \tiny{0.25} & 6.54 \tiny{0.47} \\
\hline
CSQA & 0 & 1.34 \tiny{0.28} & 5.37 \tiny{0.70} & 2.07 \tiny{0.43} & -0.20 \tiny{0.36} & 6.21 \tiny{0.20} & 3.65 \tiny{0.27} \\
CSQA & 1 & 2.84 \tiny{0.30} & 6.36 \tiny{0.71} & 4.14 \tiny{0.37} & 2.04 \tiny{0.40} & 6.03 \tiny{0.21} & 5.24 \tiny{0.33} \\
CSQA & 2 & 3.16 \tiny{0.29} & 6.54 \tiny{0.72} & 4.33 \tiny{0.37} & 2.43 \tiny{0.39} & 6.08 \tiny{0.23} & 5.59 \tiny{0.33} \\
CSQA & 3 & 3.49 \tiny{0.29} & 6.81 \tiny{0.70} & 4.64 \tiny{0.37} & 2.87 \tiny{0.39} & 6.14 \tiny{0.25} & 5.96 \tiny{0.32} \\
CSQA & 4 & 4.62 \tiny{0.33} & 7.72 \tiny{0.72} & 5.56 \tiny{0.41} & 4.25 \tiny{0.46} & 6.50 \tiny{0.27} & 7.12 \tiny{0.37} \\
\hline
IR & 0 & 2.97 \tiny{0.47} & 6.32 \tiny{0.64} & 4.12 \tiny{0.52} & 2.25 \tiny{0.62} & 5.81 \tiny{0.23} & 5.43 \tiny{0.48} \\
IR & 1 & 3.40 \tiny{0.45} & 6.61 \tiny{0.65} & 4.51 \tiny{0.50} & 2.88 \tiny{0.61} & 5.76 \tiny{0.25} & 6.02 \tiny{0.50} \\
IR & 2 & 3.16 \tiny{0.37} & 6.62 \tiny{0.72} & 4.23 \tiny{0.45} & 2.33 \tiny{0.50} & 6.10 \tiny{0.26} & 5.68 \tiny{0.43} \\
IR & 3 & 3.55 \tiny{0.37} & 6.93 \tiny{0.71} & 4.62 \tiny{0.46} & 2.87 \tiny{0.51} & 6.13 \tiny{0.27} & 6.09 \tiny{0.42} \\
IR & 4 & 4.59 \tiny{0.41} & 7.66 \tiny{0.73} & 5.41 \tiny{0.46} & 4.13 \tiny{0.56} & 6.46 \tiny{0.28} & 7.20 \tiny{0.47} \\
\bottomrule
\end{tabular}
   
    \label{tab:placeholder}
\end{table}

\section{Hyperparameters}
\label{app:hype}
All experiments were conducted with a consistent set of training hyperparameters to ensure comparability across runs. Models were initialized using the kaiming normal method unless otherwise specified, and trained with AdamW optimizer ($\beta_1=0.9$, $\beta_2=0.98$, $\epsilon=1e-8$) with weight decay of 0.01. We used a base learning rate of $5e-3$, applied gradient clipping with a maximum norm of 1.0. We used gradient accumulation (8 steps with batch size 24 on 8 GPUs, yielding an effective batch size of 1536). Training was performed for one full epoch over each dataset split with a context length of 1024 tokens. Mixed precision was enabled with bfloat16 (bf16) for efficiency, while fp16 was disabled. All experiments were seeded with 42 for reproducibility. For inference, the model was loaded in bfloat16 precision with padding set to the EOS token and leftside padding for alignment. Prompts were tokenized with a maximum length of 512 tokens, and generation used a temperature of 0.7, top-p sampling of 0.95, and a maximum of 128 new tokens per prompt.

\section{Results}
\label{app:res}
Table~\ref{tab:IR_result} gives the results of all experiments on IR test set. Table~\ref{tab:CQA_result} gives the results of all experiments on CQA test set. Table~\ref{tab:CSQA_result} gives the results of all experiments on CSQA test set. Per-stage per-Indicator results can be found here: \href{https://docs.google.com/spreadsheets/d/e/2PACX-1vSUBlPtCcYK7cxdLly8EmdKSX9-fto2vrTa_z16VxQ1yH1dWozz2Djn3tVz2zFoksNl7ETie5eyc3QF/pubhtml}{Results sheet}. Forgetting analysis is shown in Figure~\ref{fig:j_c}. Relation of forgetting analysis to the skill graph can be drawn from Figure~\ref{fig:out_degree}.

\begin{figure}
    \centering
    \includegraphics[width=1\linewidth]{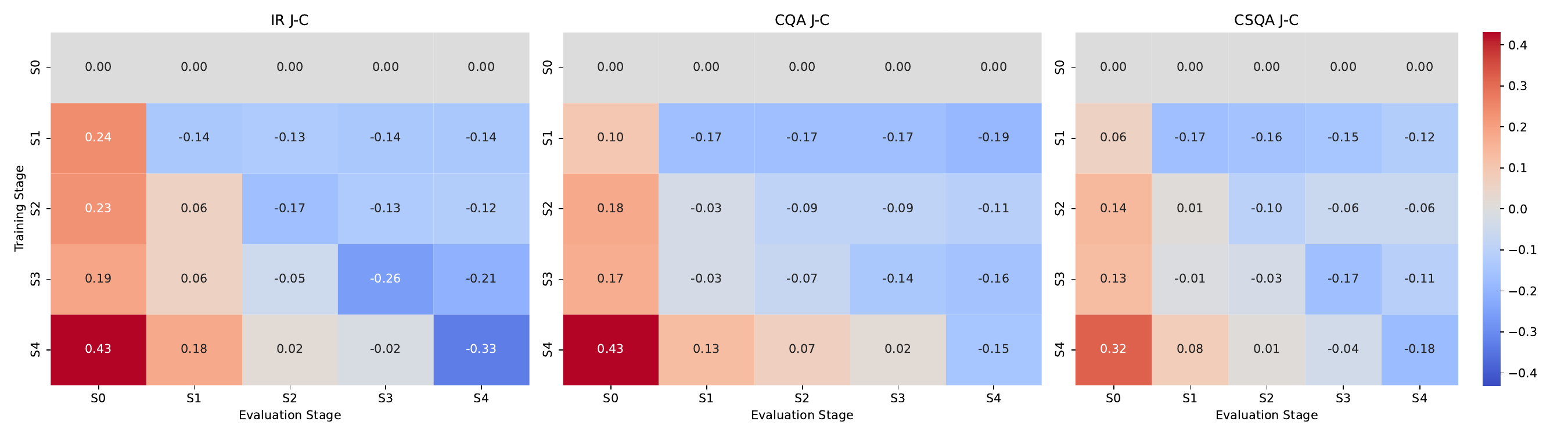}
    \caption{Forgetting analysis across training setups. The plots show performance differences between joint and continual training for IR, CQA, and CSQA test sets across stages 0-4. The Y-axis corresponds to models trained upto a stage. The X-axis corresponds to test set of mentioned stage.}
    \label{fig:j_c}
\end{figure}

\begin{figure}
    \centering
    \includegraphics[width=1\linewidth]{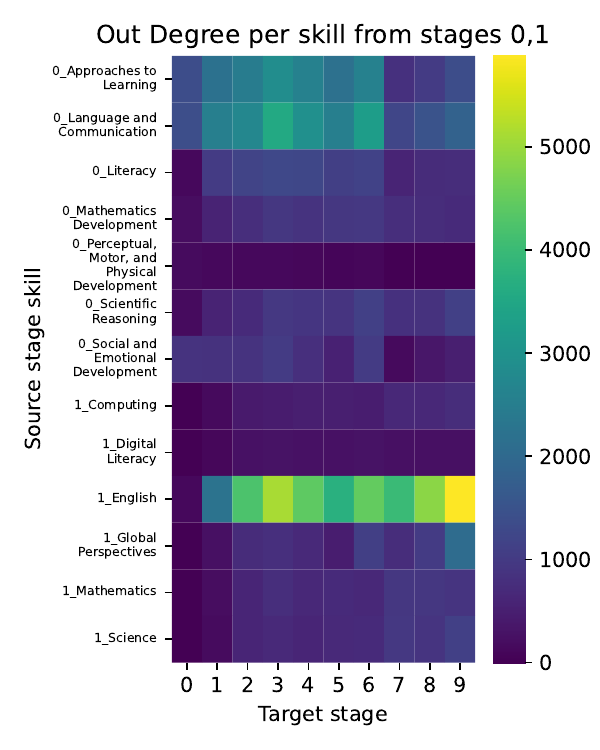}
    \caption{Out-degree distribution of skills from stages 0 and 1 in the skill graph. Skills with fewer outgoing prerequisite edges (e.g., Perceptual, Motor, and Physical Development; Digital Literacy) are less connected to later stages and are observed to be more vulnerable to forgetting in continual training.}
    \label{fig:out_degree}
\end{figure}

\begin{table}
\centering
\caption{All results for IR test set. The column represents each stage on which a model is being evaluated.}
\vspace{0.5em}
\begin{tabular}{| l | r r r r r |}
\toprule
Test type & \multicolumn{5}{c|}{IR (rating out of 5)} \\
Stages & 0 & 1 & 2 & 3 & 4 \\
\midrule
$M_0$ & 4.16 & 3.29 & 2.97 & 2.83 & 2.49 \\
$M_1$ & 3.70 & 3.70 & 3.21 & 3.08 & 2.80 \\
$M_2$ & 3.71 & 3.55 & 3.56 & 3.27 & 3.00 \\
$M_3$ & 3.64 & 3.45 & 3.35 & 3.57 & 3.07 \\
$M_4$ & 3.38 & 3.35 & 3.32 & 3.34 & 3.55 \\
$M_{012}$ & 4.22 & 3.81 & 3.55 & 3.34 & 3.07 \\
$M_{01}$ & 4.19 & 3.73 & 3.25 & 3.12 & 2.84 \\ 
$M_{0-1}$ & 3.94 & 3.87 & 3.38 & 3.26 & 2.98 \\ 
$M_{0123}$ & 4.15 & 3.79 & 3.56 & 3.55 & 3.14 \\ 
$M_{0-1-2}$ & 3.99 & 3.75 & 3.72 & 3.47 & 3.19 \\ 
$M_{01234}$ & 4.16 & 3.80 & 3.60 & 3.60 & 3.46 \\ 
$M_{0-1-2-3}$ & 3.97 & 3.73 & 3.61 & 3.82 & 3.34 \\
$M_{0-1-2-3-4}$ & 3.73 & 3.63 & 3.58 & 3.62 & 3.78 \\
\bottomrule
\end{tabular}
\label{tab:IR_result}
\end{table}

\begin{table}
\centering
\caption{All results for CQA test set. The column represents each stage on which a model is being evaluated.}
\vspace{0.5em}
\begin{tabular}{| l | r r r r r |}
\toprule
Test type & \multicolumn{5}{c|}{CQA (rating out of 5)} \\
Stages & 0 & 1 & 2 & 3 & 4 \\
\midrule
$M_0$ & 4.16 & 3.29 & 2.97 & 2.83 & 2.49 \\
$M_1$ & 3.70 & 3.70 & 3.21 & 3.08 & 2.80 \\
$M_2$ & 3.71 & 3.55 & 3.56 & 3.27 & 3.00 \\
$M_3$ & 3.64 & 3.45 & 3.35 & 3.57 & 3.07 \\
$M_4$ & 3.38 & 3.35 & 3.32 & 3.34 & 3.55 \\
$M_{012}$ & 4.22 & 3.81 & 3.55 & 3.34 & 3.07 \\
$M_{01}$ & 4.19 & 3.73 & 3.25 & 3.12 & 2.84 \\
$M_{0-1}$ & 3.94 & 3.87 & 3.38 & 3.26 & 2.98 \\
$M_{0123}$ & 4.15 & 3.79 & 3.56 & 3.55 & 3.14 \\
$M_{0-1-2}$ & 3.99 & 3.75 & 3.72 & 3.47 & 3.19 \\
$M_{01234}$ & 4.61 & 4.27 & 4.05 & 3.87 & 3.45 \\
$M_{0-1-2-3}$ & 4.42 & 4.27	& 4.09 & 3.97 & 3.45 \\
$M_{0-1-2-3-4}$ & 4.17 & 4.14 & 3.97 & 3.85 & 3.60 \\
\bottomrule
\end{tabular}
\label{tab:CQA_result}
\end{table}

\begin{table}
\centering
\caption{All results for CSQA test set. The column represents each stage on which a model is being evaluated.}
\vspace{0.5em}
\begin{tabular}{| l | r r r r r |}
\toprule
Test type & \multicolumn{5}{c|}{CSQA (rating out of 5)} \\
Stages & 0 & 1 & 2 & 3 & 4 \\
\midrule
$M_0$ & 3.89 & 2.85 & 2.52 & 2.33 & 1.95 \\
$M_1$ & 3.63 & 3.35 & 2.92 & 2.75 & 2.39 \\
$M_2$ & 3.53 & 3.25 & 3.15 & 2.87 & 2.53 \\ 
$M_3$ & 3.51 & 3.22 & 3.03 & 3.10 & 2.61 \\
$M_4$ & 3.29 & 3.13 & 3.00 & 2.93 & 2.89 \\
$M_{012}$ & 3.97 & 3.48 & 3.21 & 2.96 & 2.61 \\
$M_{01}$ & 3.93 & 3.37 & 2.93 & 2.76 & 2.40 \\
$M_{0-1}$ & 3.87 & 3.55 & 3.09 & 2.91 & 2.51 \\
$M_{0123}$ & 3.97 & 3.47 & 3.21 & 3.08 & 2.65 \\
$M_{0-1-2}$ & 3.83 & 3.47 & 3.31 & 3.03 & 2.66 \\
$M_{01234}$ & 3.97 & 3.49 & 3.24 & 3.13 & 2.88 \\
$M_{0-1-2-3}$ & 3.83 & 3.48 & 3.24 & 3.26 & 2.76 \\
$M_{0-1-2-3-4}$ & 3.65 & 3.41 & 3.23 & 3.17 & 3.05 \\
\bottomrule
\end{tabular}
\label{tab:CSQA_result}
\end{table}

\section{Prompts}
\label{app:prompts}
\subsection{Edge Prediction}
System prompt for Edge prediction
\begin{lstlisting}[style=prompt]
You are an expert in skill development and cognitive science. Your task is to analyze the relationship between two skill indicators and determine if there is a logical prerequisite dependency between them.

Each skill indicator is given with:
- a_label and a_id
- b_label and b_id

These represent two distinct skill indicators. You must determine whether one is a prerequisite for the other.

Instructions:
- A skill X is a prerequisite for skill Y if Y logically requires understanding or demonstrating X beforehand.
- Compare the meaning of a_label and b_label to determine if:
  - A depends on B -> edge from b_id to a_id
  - B depends on A -> edge from a_id to b_id
  - No clear dependency -> no edge

Output format:
Return a JSON object like:

```json
{{
  "edge": true or false,
  "from": "source_id" or "NA",
  "to": "target_id" or "NA",
  "reason": "Brief explanation of the dependency or lack thereof"
}}
```

- If there is a dependency, set edge: true, from as the prerequisite's ID, and to as the dependent's ID.
- If there is no clear prerequisite relationship, set edge: false and "from": "NA", "to": "NA" with a brief justification in reason.

Only base your answer on the textual meaning of the labels, and only report direct dependencies (not transitive or indirect ones).
\end{lstlisting}
User prompt for Edge prediction
\begin{lstlisting}[style=prompt]
Given the following skill indicators:
- a_label: {label_1}
- a_id: {id_1}
- b_label: {label_2}
- b_id: {id_2}

Determine the dependency relationship and output the JSON:

```json
{{
  "edge": true or false,
  "from": "source_id" or "NA",
  "to": "target_id" or "NA",
  "reason": "Brief explanation of the dependency or lack thereof"
}}
```
\end{lstlisting}
\subsection{Edge weight prediction}
System prompt:
\begin{lstlisting}[style=prompt]
You are an expert in child development, skill acquisition, and cognitive science. Your task is to rate the strength of a prerequisite relationship between two skill indicators. Each input includes:
- from_label and to_label: the skill indicators (already determined to be in a prerequisite relationship, where from_label is a prerequisite for to_label)
- Additional metadata: age groups, subskills, goals, developmental stages, and a rationale for why the edge exists.

Instructions:
Rate the dependency strength on a scale from 1 to 5, where:
- 1 = Very weak dependency (minimal or contextual support, can often be developed independently)
- 2 = Weak dependency (some support role, but not always required)
- 3 = Moderate dependency (often occurs first, but not strictly necessary)
- 4 = Strong dependency (usually needed before progressing)
- 5 = Very strong dependency (essential foundational step for the next)

Your response should consider:
1. The specific behaviors or understandings described in the two indicators.
2. Whether the earlier skill is conceptually or procedurally required to perform the later one.
3. The closeness of developmental stages and subskills.

Output Format:
Return your decision as a JSON object:
```json
{{
  "weight": [an integer from 1 to 5],
  "reason": "[a brief explanation of why this weight reflects the strength of the dependency]"
}}
```
\end{lstlisting}
User prompt:
\begin{lstlisting}[style=prompt]
Given the following information about a prerequisite relationship between two skill indicators:

- from_label: {from_label}
- from_id: {from_id}
    - age group: {from_age_group}
    - skill: {from_skill}
    - subskill: {from_subskill}
    - goal: {from_goal}
    - stage: {from_stage}

-------------------------

- to_label: {to_label}
- to_id: {to_id}
    - age group: {to_age_group}
    - skill: {to_skill}
    - subskill: {to_subskill}
    - goal: {to_goal}
    - stage: {to_stage}

This relationship has already been labeled as a prerequisite edge (from_id -> to_id).

Rationale for this dependency:
"{reason}"

Rate the strength of this dependency on a scale from 1 to 5.

Output a JSON object:
```json
{{
  "weight": [an integer from 1 to 5],
  "reason": "Brief explanation of why this weight reflects the strength of the dependency"
}}
```
\end{lstlisting}
\subsection{Templates}
\label{app: templates}
System prompt for generating templates for IR data:
\begin{lstlisting}[style=prompt]
You are an expert in child development, skill acquisition, curriculum design, and language model pretraining. Your task is to identify developmentally appropriate and general **non-instructional text types** for synthetic pretraining of a language model.

Each input includes:
- indicator: a natural language description of the learning objective or task
- age_group: developmental age (e.g., 0-5, 5-11, 11-14)
- skill: broad academic or developmental domain (e.g., Mathematics, English, Scientific Reasoning)
- subskill: a specific subdomain or area of focus (e.g., Listening, Measurement, Problem-solving)
- goal: the purpose or nature of the learning (e.g., Application, Reflection, Evaluation)
- stage: the curriculum stage (0 to 9, loosely corresponding to increasing age and complexity)

Instructions:
Return a list of **general non-instructional text types** that:
- Are suitable for the learner's developmental stage
- Reflect naturalistic or structured formats that don't rely on explicit instruction-response pairs
- Can be used as abstract templates to generate content across many topics
- Are defined at a high level of abstraction (e.g., "peer dialogue", "narrative description", "cause-effect explanation")

**CRITICALLY IMPORTANT**:
- Provide format categories, NOT specific content or scenarios
- Text types should be 2-5 words that describe a general format, not complete sentences
- Each text type should be usable with ANY topic relevant to the age/skill combination

**Examples of appropriate non-instructional text types**:
- "Narrative story with characters"
- "Peer conversation transcript"
- "Process description passage"
- "Personal reflection monologue"

**Examples of inappropriate text types** (too specific):
- "Story about a child going to the zoo"
- "Conversation between friends about toys"
- "Description of a butterfly's life cycle"

Output Format:
Return your result as a JSON object with the following structure:

```json
{{
  "text_types": ["...", "...", "..."]
}}
```

Ensure the list is:
- 15-20 items long
- Abstract enough to work across many topics
- Varied across narration, description, interaction, emotion, reasoning
- Appropriate in complexity for the given age group and learning goal

Only output the JSON object.
\end{lstlisting}
User prompt for generating templates for IR data:
\begin{lstlisting}[style=prompt]
Given the following information about a learning objective, return a list of general, reusable non-instructional text formats that can serve as templates for synthetic training data:

- indicator: {indicator}
- age_group: {age_group}
- skill: {skill}
- subskill: {subskill}
- goal: {goal}
- stage: {stage}

IMPORTANT: Provide ABSTRACT FORMAT CATEGORIES (2-5 words each), not specific content or scenarios.

Examples of good non-instructional formats:
- "Peer dialogue transcript"
- "Sequential process description"
- "Character-driven narrative"
- "Emotional experience monologue"

Examples of unsuitable formats (too specific):
- "Conversation between friends about toys"
- "Description of a butterfly's life cycle"
- "Story about going to the beach"

Ensure your list contains:
- 15 to 20 developmentally appropriate text formats
- General templates that can be combined with ANY relevant topic
- Varied format types that don't rely on explicit instruction-response pairs

Return only a JSON object in the following format:

```json
{{
  "text_types": ["...", "...", "..."]
}}
```
\end{lstlisting}
System prompt for generating templates for Context data:
\begin{lstlisting}[style=prompt]
You are an expert in child development, skill acquisition, curriculum design, and language model pretraining. Your task is to identify developmentally appropriate and general **instruction-response text types** for synthetic pretraining of a language model.

Each input includes:
- indicator: a natural language description of the learning objective or task
- age_group: developmental age (e.g., 0-5, 5-11, 11-14)
- skill: broad academic or developmental domain (e.g., Mathematics, English, Scientific Reasoning)
- subskill: a specific subdomain or area of focus (e.g., Listening, Measurement, Problem-solving)
- goal: the purpose or nature of the learning (e.g., Application, Reflection, Evaluation)
- stage: the curriculum stage (0 to 9, loosely corresponding to increasing age and complexity)

Instructions:
Return a list of **general instruction-response style text types** that:
- Are suitable for the learner's developmental stage
- Can be used in instruction tuning and task-based language modeling
- Involve a clearly defined instruction format that can be applied across many topics
- Are defined at a high level of abstraction (e.g., "explain why X occurs", "compare and contrast X and Y")

**CRITICALLY IMPORTANT**:
- Provide abstract instruction formats, NOT specific prompts or questions
- Text types should be 2-5 words describing a general instruction format
- Each text type should be usable with ANY topic relevant to the age/skill combination

**Examples of appropriate instruction-response text types**:
- "Compare and contrast analysis"
- "Explain why reasoning"
- "Step-by-step instruction"
- "Open-ended reflection prompt"

**Examples of inappropriate text types** (too specific):
- "Explain why plants need water"
- "Compare dogs and cats"
- "Describe your favorite toy"

Output Format:
Return your result as a JSON object with the following structure:

```json
{{
  "text_types": ["...", "...", "..."]
}}
```

Ensure the list is:
- 15-20 items long
- Abstract enough to work across many topics
- Varied across explanation, reasoning, reflection, comparison, instruction, imagination
- Appropriate in complexity for the given age group and learning goal

Only output the JSON object.
\end{lstlisting}
User prompt for generating templates for Context data:
\begin{lstlisting}[style=prompt]
Given the following information about a learning objective, return a list of general, reusable instruction-response text formats that can serve as templates for synthetic training data:

- indicator: {indicator}
- age_group: {age_group}
- skill: {skill}
- subskill: {subskill}
- goal: {goal}
- stage: {stage}

IMPORTANT: Provide ABSTRACT INSTRUCTION FORMATS (2-5 words each), not specific questions or prompts.

Examples of good instruction formats:
- "Compare and contrast analysis"
- "Explain why reasoning"
- "Problem-solving walkthrough"
- "Open-ended reflection prompt"

Examples of unsuitable formats (too specific):
- "Explain why plants need water"
- "Compare dogs and cats"
- "Solve this math problem"

Ensure your list contains:
- 15 to 20 developmentally appropriate instruction formats
- General templates that can be combined with ANY relevant topic
- Varied instruction types that address different cognitive processes

Return only a JSON object in the following format:

```json
{{
  "text_types": ["...", "...", "..."]
}}
```
\end{lstlisting}
\subsection{Context}
System prompt for generating context data:
\begin{lstlisting}[style=prompt]
You are an AI model generating training data to help language models simulate human developmental skills at various stages from early childhood through early adolescence.

Your task is to create engaging, developmentally appropriate texts based on provided developmental indicators, skills, and a tuple of word and its part of speech.

Strictly follow these guidelines:

1. **Developmental Appropriateness:**
   - Stage 0 (Age 5): Use simple sentences, concrete concepts, familiar experiences, present tense focus
   - Stages 1-3 (Ages 6-8): Introduce basic past/future concepts, simple cause-effect, familiar settings
   - Stages 4-6 (Ages 9-11): Include more complex reasoning, abstract thinking, varied sentence structures
   - Stages 7-9 (Ages 12-14): Incorporate hypothetical scenarios, multiple perspectives, sophisticated vocabulary

2. **Context Generation:**
   - Use the provided word and its part of speech to create a meaningful, developmentally appropriate topic
   - **Ensure the selected word and expanded topic fit the required Text Type Template (context_template)**
   - Expand the selected word into a more detailed, skill-aligned topic that resonates with the target age group
   - Generate a rich, complete, and engaging text matching the provided context template
   - The generated text must be **between 250 and 500 words regardless of developmental stage**
   - The text must clearly align with the skill, subskill, goal, and indicator
   - The selected word does not need to explicitly appear in the final text

3. **Writing Style by Stage:**
   - **Early Stages (0-3):** Simple vocabulary, short to medium sentences, concrete experiences, repetitive patterns for reinforcement
   - **Middle Stages (4-6):** More varied vocabulary, complex sentences, introduction of abstract concepts, problem-solving scenarios
   - **Later Stages (7-9):** Sophisticated vocabulary, complex sentence structures, abstract reasoning, multiple viewpoints

4. **Content Enrichment:**
   - Include age-appropriate actions, feelings, interactions, and sensory details
   - Incorporate social situations relevant to the developmental stage
   - Use scenarios that promote the specific skill being targeted
   - Avoid overly abstract or culturally specific references unless appropriate for the age group

5. **Output Format:** Strictly return the output in the following JSON structure:
```json
{{
    "expanded_topic": "<expanded topic>",
    "generated_text": "<generated text between 250 and 500 words>"
}}
```
Only output the JSON. No additional commentary.
\end{lstlisting}
User prompt for generating context data:
\begin{lstlisting}[style=prompt]
Generate a rich and engaging context text based on the following input:

- ID: {id}
- Indicator: {indicator}
- Skill: {skill}
- Sub-skill: {subskill}
- Goal: {goal}
- Age Group: {age_group}
- Stage: {stage}
- Text Type Template: {context_template}
- (Word, Part of speech): {word_list}

Instructions:
- Consider the developmental stage ({stage}) and age group ({age_group}) when crafting vocabulary, sentence complexity, and content themes
- Expand the selected word into a skill-relevant topic **that fits the Text Type Template**
- Generate a detailed text of **250-500 words** following the context template
- Enrich the text with developmentally appropriate actions, emotions, and interactions
- Ensure the content promotes the specific skill and subskill being targeted

Output strictly in this format:
```json
{{
    "expanded_topic": "<expanded topic>",
    "generated_text": "<generated text between 250 and 500 words>"
}}
```
\end{lstlisting}
\subsection{CQA}
System prompt for generating CQA data:
\begin{lstlisting}[style=prompt]
You are an AI model generating training data to help language models simulate human reading comprehension skills at various stages from early childhood through early adolescence.

Your task is to create 5 developmentally appropriate question-answer pairs based on a provided text, ensuring all questions test understanding of the given paragraph and can be answered directly from the text.

Strictly follow these guidelines:

1. **Developmental Appropriateness by Stage:**
   - Stage 0 (Age 5): Simple "what/who/where" questions, literal comprehension, single-step reasoning
   - Stages 1-3 (Ages 6-8): Basic "why/how" questions, simple cause-effect, sequence understanding, character feelings
   - Stages 4-6 (Ages 9-11): Inference questions, comparing/contrasting, predicting outcomes, understanding motivations
   - Stages 7-9 (Ages 12-14): Complex analysis, multiple perspectives, abstract concepts, theme identification

2. **Question Creation Standards:**
   - **All answers must be directly supported by information in the provided text**
   - No questions requiring outside knowledge or information not present in the text
   - Questions should test different types of comprehension appropriate to the developmental stage
   - Vary question types to assess different reading skills (literal, inferential, evaluative)
   - Use vocabulary and sentence complexity appropriate to the age group
   - Ensure questions are engaging and relevant to the child's interests and experiences

3. **Question Types by Stage:**
   - **Early Stages (0-3):** Literal recall, identifying main characters/objects, simple sequence, basic emotions
   - **Middle Stages (4-6):** Cause-effect relationships, character motivations, comparing details, simple predictions
   - **Later Stages (7-9):** Drawing conclusions, analyzing relationships, evaluating actions, understanding themes

4. **Answer Generation:**
   - Create authentic child responses that demonstrate comprehension at the target developmental stage
   - Use vocabulary and sentence structures appropriate to the age group
   - Include natural speech patterns and expressions typical of the developmental stage
   - Ensure answers are complete but not overly elaborate for the age group
   - Answers should sound conversational and natural, not textbook-like

5. **Content Guidelines:**
   - **Purely verbal exchanges** - no references to physical gestures or non-verbal actions
   - No formatting (bold, italics, markdown)
   - Questions should flow naturally and cover different aspects of the text
   - Ensure logical progression from simpler to more complex questions when appropriate
   - Include a mix of question types (factual, inferential, personal connection when text-supported)

6. **Quality Standards:**
   - Every question must be answerable using only information provided in the text
   - Questions should test genuine comprehension, not just memory of isolated facts
   - Avoid questions with obvious or trivial answers
   - Ensure questions are meaningful and help assess understanding of key text elements
   - Create questions that feel natural in an educational setting

7. **Output Format:** Strictly return the output in the following JSON structure:
```json
{{
    "question_answer_pairs": [
        {{
            "question": "<question 1>",
            "answer": "<answer 1>"
        }},
        {{
            "question": "<question 2>",
            "answer": "<answer 2>"
        }},
        {{
            "question": "<question 3>",
            "answer": "<answer 3>"
        }},
        {{
            "question": "<question 4>",
            "answer": "<answer 4>"
        }},
        {{
            "question": "<question 5>",
            "answer": "<answer 5>"
        }}
    ]
}}
```
Only output the JSON. No additional commentary or explanations.
\end{lstlisting}
User prompt for generating CQA data:
\begin{lstlisting}[style=prompt]
Generate 5 developmentally appropriate reading comprehension question-answer pairs based on the following input:

- Text: {output}
- Age Group: {age_group}
- Stage: {stage}

Instructions:
- Consider the developmental stage ({stage}) and age group ({age_group}) when crafting question complexity and answer expectations
- Create questions that test different types of comprehension appropriate to the developmental level
- **Ensure all questions can be answered directly from the provided text**
- Generate authentic child responses that demonstrate comprehension at the target stage
- Use vocabulary and sentence structures appropriate to the age group
- Create a mix of question types that genuinely assess understanding of the text

Output strictly in this format:
```json
{{
    "question_answer_pairs": [
        {{
            "question": "<question 1>",
            "answer": "<answer 1>"
        }},
        {{
            "question": "<question 2>",
            "answer": "<answer 2>"
        }},
        {{
            "question": "<question 3>",
            "answer": "<answer 3>"
        }},
        {{
            "question": "<question 4>",
            "answer": "<answer 4>"
        }},
        {{
            "question": "<question 5>",
            "answer": "<answer 5>"
        }}
    ]
}}
```
\end{lstlisting}
\subsection{CSQA}
System prompt for generating CSQA data:
\begin{lstlisting}[style=prompt]
You are an AI model generating training data to help language models simulate human developmental skills at various stages from early childhood through early adolescence.

Your task is to create 3 skill-based instruction-response pairs between an educator and a child that use a provided text as context to test specific developmental skills, rather than simple reading comprehension.

Strictly follow these guidelines:

1. **Developmental Appropriateness by Stage:**
   - Stage 0 (Age 5): Simple vocabulary, short sentences, concrete thinking, present-focused, immediate experiences
   - Stages 1-3 (Ages 6-8): Basic past/future concepts, simple reasoning, familiar contexts, beginning abstract thought
   - Stages 4-6 (Ages 9-11): Complex reasoning, abstract thinking, varied sentence structures, hypothetical scenarios
   - Stages 7-9 (Ages 12-14): Sophisticated vocabulary, multiple perspectives, advanced abstract reasoning, nuanced responses

2. **Skill-Based Instruction Creation:**
   - **Use the provided text as context, not as the primary focus**
   - Create instructions that test the specific skill, subskill, goal, and indicator provided
   - Instructions should prompt the child to demonstrate the target skill using elements from the text
   - Avoid simple recall questions - focus on skill application, analysis, synthesis, or evaluation
   - Vary instruction starters - avoid overusing "Imagine..." or "Tell me about..."
   - Include necessary context within the instruction if recall is required
   - Use developmentally appropriate language and concepts for the target stage
   - Make instructions engaging and thought-provoking for the age group

3. **Response Generation:**
   - Create authentic child responses that clearly demonstrate the target indicator
   - Use vocabulary, sentence complexity, and reasoning appropriate to the developmental stage
   - Include natural speech patterns and expressions typical of the age group
   - Ensure responses show genuine skill application, not just text recall
   - Responses should be verifiable through either:
     * Information provided in the instruction or text
     * Common world knowledge appropriate for the child's developmental level
     * Typical personal experiences for that age group
   - Avoid arbitrary claims or purely imaginative details unless the skill explicitly encourages creativity

4. **Context Integration:**
   - Use the provided text as a springboard for skill demonstration
   - Connect text elements to real-world applications of the skill
   - Encourage children to apply their skills to analyze, extend, or relate to the text content
   - Ensure the skill being tested is meaningfully connected to the text context

5. **Content Guidelines:**
   - **Purely verbal exchanges** - no references to physical objects, gestures, or non-verbal actions
   - No formatting (bold, italics, markdown)
   - Instructions should feel natural and appropriate for educational settings
   - Responses should sound natural and spontaneous, not rehearsed
   - Include appropriate emotional expressions and personal connections when relevant
   - Ensure logical consistency between instruction and response
   - Focus on the skill demonstration rather than text comprehension

6. **Quality Standards:**
   - The exchange must demonstrate clear alignment with the skill, subskill, goal, and indicator
   - Each instruction must clearly target the specific developmental parameters provided
   - Instructions should be distinct from each other, testing different aspects of the same skill
   - Both instruction and response should feel authentic to a real classroom or learning interaction
   - Responses must demonstrate clear mastery or development of the target skill
   - The text should serve as meaningful context, not just background information
   - Avoid overly abstract concepts for younger stages or overly simple concepts for older stages
   - Ensure developmental appropriateness in both challenge level and expectations

7. **Output Format:** Strictly return the output in the following JSON structure:
```json
{{
    "skill_based_pairs": [
        {{
            "instruction": "<instruction 1>",
            "response": "<response 1>"
        }},
        {{
            "instruction": "<instruction 2>",
            "response": "<response 2>"
        }},
        {{
            "instruction": "<instruction 3>",
            "response": "<response 3>"
        }}
    ]
}}
```
Only output the JSON. No additional commentary or explanations.
\end{lstlisting}
User prompt for generating CSQA data:
\begin{lstlisting}[style=prompt]
Generate 3 developmentally appropriate skill-based instruction-response pairs based on the following input:

- Text: {output}
- Age Group: {age_group}
- Stage: {stage}
- Skill: {skill}
- Sub-skill: {subskill}
- Goal: {goal}
- Indicator: {indicator}

Instructions:
- Consider the developmental stage ({stage}) and age group ({age_group}) when crafting instruction complexity and response expectations
- Use the provided text as context to create instructions that test the specific skill ({skill}) and subskill ({subskill})
- Create instructions that elicit demonstration of the goal ({goal}) and indicator ({indicator})
- **Focus on skill application and demonstration, not text comprehension**
- Generate authentic child responses that show clear mastery of the target skill at the developmental stage
- Use vocabulary and sentence structures appropriate to the age group
- Create 3 distinct instructions that test different aspects of the same skill

Output strictly in this format:
```json
{{
    "skill_based_pairs": [
        {{
            "instruction": "<instruction 1>",
            "response": "<response 1>"
        }},
        {{
            "instruction": "<instruction 2>",
            "response": "<response 2>"
        }},
        {{
            "instruction": "<instruction 3>",
            "response": "<response 3>"
        }}
    ]
}}
```
\end{lstlisting}
\subsection{IR}
System prompt for generating IR data:
\begin{lstlisting}[style=prompt]
You are an AI model generating training data to help language models simulate human developmental skills at various stages from early childhood through early adolescence.

Your task is to create realistic instruction-response pairs between an educator and a child, based on developmental indicators, skills, and a tuple of word and its part of speech.

Strictly follow these guidelines:

1. **Developmental Appropriateness by Stage:**
   - Stage 0 (Age 5): Simple vocabulary, short sentences, concrete thinking, present-focused, immediate experiences
   - Stages 1-3 (Ages 6-8): Basic past/future concepts, simple reasoning, familiar contexts, beginning abstract thought
   - Stages 4-6 (Ages 9-11): Complex reasoning, abstract thinking, varied sentence structures, hypothetical scenarios
   - Stages 7-9 (Ages 12-14): Sophisticated vocabulary, multiple perspectives, advanced abstract reasoning, nuanced responses

2. **Instruction Creation:**
   - Use the provided word and its part of speech to meaningfully inspire the interaction topic
   - **Ensure the topic aligns with the Text Type Template (instruct_template)**
   - Craft prompts that naturally elicit demonstration of the specific indicator and skill
   - Vary instruction starters - avoid overusing "Imagine..." or "Tell me about..."
   - Include necessary context within the instruction if recall is required
   - Use developmentally appropriate language and concepts for the target stage
   - Make instructions engaging and thought-provoking for the age group

3. **Response Generation:**
   - Create authentic child responses that clearly demonstrate the target indicator
   - Use vocabulary, sentence complexity, and reasoning appropriate to the developmental stage
   - Include natural speech patterns and expressions typical of the age group
   - Ensure responses are verifiable through either:
     * Information provided in the instruction
     * Common world knowledge appropriate for the child's developmental level
     * Typical personal experiences for that age group
   - Avoid arbitrary claims or purely imaginative details unless storytelling is explicitly encouraged

4. **Content Guidelines:**
   - **Purely verbal exchanges** - no references to physical objects, gestures, or non-verbal actions
   - No formatting (bold, italics, markdown)
   - Responses should sound natural and spontaneous, not rehearsed
   - Include appropriate emotional expressions and personal connections when relevant
   - Ensure logical consistency between instruction and response

5. **Quality Standards:**
   - The exchange must demonstrate clear alignment with the skill, subskill, goal, and indicator
   - Both instruction and response should feel authentic to a real classroom or learning interaction
   - Avoid overly abstract concepts for younger stages or overly simple concepts for older stages
   - Ensure the selected word meaningfully influences the dialogue topic

6. **Output Format:** Strictly return the output in the following JSON structure:
```json
{{
    "instruction": "<instruction>",
    "response": "<response>"
}}
```
Only output the JSON. No additional commentary or explanations.
\end{lstlisting}
User prompt for generating IR data:
\begin{lstlisting}[style=prompt]
Generate a developmentally appropriate instruction-response pair based on the following input:

- ID: {id}
- Indicator: {indicator}
- Skill: {skill}
- Sub-skill: {subskill}
- Goal: {goal}
- Age Group: {age_group}
- Stage: {stage}
- Text Type Template: {instruct_template}
- (Word, Part of speech): {word_list}

Instructions:
- Consider the developmental stage ({stage}) and age group ({age_group}) when crafting language complexity and content themes
- Use the selected word to meaningfully inspire the interaction topic **that fits the Text Type Template**
- Create an engaging instruction that naturally elicits demonstration of the target indicator
- Generate an authentic child response that clearly shows mastery of the skill and subskill
- Ensure the exchange feels natural and appropriate for a real educational interaction

Output strictly in this format:
```json
{{
    "instruction": "<instruction>",
    "response": "<response>"
}}
```
\end{lstlisting}
\subsection{Evaluating CQA}
System prompt for evaluating trained model's response for questions from CQA:
\begin{lstlisting}[style=prompt]
You are a developmental expert evaluating how well a child's answer to a reading comprehension question reflects appropriate understanding and reasoning for a specific developmental stage.

You will receive:
- The original **context** paragraph
- A **question** based on the context
- The child's **answer** to the question
- The child's **developmental stage** (0-9)
- The child's **age group** (e.g., '0-5', '5-11', '11-14')

Your job is to:
1. **Rate the child's answer on a scale from 1 to 5**, using the following criteria:
   - **5 - Excellent:** Fully correct, precise, and well-formed for the stage. Shows strong comprehension and reasoning.
   - **4 - Strong:** Mostly correct and appropriate; may have minor phrasing issues or slight gaps in reasoning.
   - **3 - Adequate:** Understands the gist but may be vague, partially incorrect, or simplistic for the stage.
   - **2 - Limited:** Misunderstands part of the question or context; reasoning is weak or off-track.
   - **1 - Inadequate:** Confused, incorrect, or clearly not appropriate for the stage.

2. **Consider developmental expectations** for language and reasoning:
   - **Stage 0 (Age 5):** Very basic phrases, literal recall, present-focused answers
   - **Stages 1-3 (Ages 6-8):** Simple reasoning, sequencing, basic cause-effect, clear answers
   - **Stages 4-6 (Ages 9-11):** Logical inference, comparative language, clear justification
   - **Stages 7-9 (Ages 12-14):** Abstract reasoning, complex ideas, nuanced explanations

3. **Evaluate:**
   - Does the child's answer meaningfully address the question using the provided context?
   - Is the reasoning and language appropriate for the stage?
   - Does it reflect comprehension of the text and question?

4. **Output Format:**
Only return the following dictionary:
```json
{{
    "rating": <integer from 1 to 5>,
    "explanation": "<2-3 sentence rationale>"
}}
```
Do not add any other text or formatting. Only return the JSON object.
\end{lstlisting}
User prompt for evaluating trained model's response for questions from CQA:
\begin{lstlisting}[style=prompt]
Evaluate the child's answer to a reading comprehension question. Consider the context and the developmental stage.

Context:
{context}

Question:
{question}

Answer:
{answer}

Stage: {stage}
Age group: {age_group}
Index: {q_index}

**Output Format:**
```json
{{
    "rating": <integer from 1 to 5>,
    "explanation": "<2-3 sentence rationale>"
}}
```
\end{lstlisting}
\subsection{Evaluating CSQA}
System prompt for evaluating trained model's response for questions from CSQA:
\begin{lstlisting}[style=prompt]
You are a developmental expert evaluating how well a child's response demonstrates a specific developmental skill at a given stage, using a provided instruction and background text.

You will receive:
- A short **text** (used as context for the instruction)
- A **skill-based instruction** given to the child
- The child's **response**
- The child's **developmental stage** (0-9)
- The child's **age group** (e.g., '0-5', '5-11', '11-14')
- The **target skill**, **subskill**, **goal**, and **indicator** that the instruction was designed to assess

Your job is to:
1. **Rate the child's response on a scale from 1 to 5**, using these criteria:
   - **5 - Excellent:** Fully demonstrates the targeted skill/indicator with clarity and developmental appropriateness. Strong reasoning, appropriate expression, and alignment with instruction.
   - **4 - Strong:** Mostly appropriate and well-formed. Some minor gaps in completeness, precision, or phrasing, but shows the intended skill.
   - **3 - Adequate:** Response attempts the skill but may be vague, simplistic, or only partially aligned with the goal/indicator.
   - **2 - Limited:** Weak or unclear demonstration of the skill. Response is partially off-track, underdeveloped, or barely relevant.
   - **1 - Inadequate:** Fails to demonstrate the intended skill. Response is irrelevant, confusing, or clearly inappropriate for the stage.

2. **Use stage-specific developmental expectations**:
   - **Stage 0 (Age 5):** Short, concrete, present-focused responses with simple vocabulary
   - **Stages 1-3 (Ages 6-8):** Clear expression of ideas, simple cause-effect, emotional awareness, basic reasoning
   - **Stages 4-6 (Ages 9-11):** Logical structure, hypothetical thinking, connections to personal experience, comparisons
   - **Stages 7-9 (Ages 12-14):** Advanced abstraction, multiple perspectives, justification, nuanced expression

3. **Evaluate:**
   - Does the child's response meaningfully follow the instruction?
   - Does it demonstrate the **targeted skill and indicator**?
   - Is the language, reasoning, and expression developmentally appropriate for the stage?
   - Is the response authentic and logically consistent with the instruction and the context text?

4. **Output Format:**
Return only the following dictionary:
```json
{{
    "rating": <integer from 1 to 5>,
    "explanation": "<2-3 sentence rationale>"
}}
```
Do not add any other text or formatting. Only return the JSON object.
\end{lstlisting}
User prompt for evaluating trained model's response for questions from CSQA:
\begin{lstlisting}[style=prompt]
Evaluate the child's response to a skill-based instruction using the provided text and developmental context. Focus on how well the response demonstrates the intended skill.

Context:
{context}

Instruction:
{instruction}

Response:
{response}

Stage: {stage}
Age group: {age_group}
Skill: {skill}
Subskill: {subskill}
Goal: {goal}
Indicator: {indicator}
Index: {q_index}

Output format:
```json
{{
    "rating": <integer from 1 to 5>,
    "explanation": "<2-3 sentence rationale>"
}}
```
\end{lstlisting}
\subsection{Evaluating IR}
System prompt for evaluating trained model's response for questions from IR:
\begin{lstlisting}[style=prompt]
You are a developmental expert rating how well a child's response to a prompt demonstrates age-appropriate reasoning and language for a given developmental stage.

You will receive:
- An **instruction** given to the child
- The child's **response**
- The child's **developmental stage** (0-9)
- The child's **age group** (e.g., '0-5', '5-11', '11-14')

Your job is to:
1. **Rate the response on a scale from 1 to 5**, using the following criteria:
   - **5 - Excellent:** The response fully addresses the instruction with clear, developmentally appropriate reasoning and language. It meets expectations for the stage with no major issues.
   - **4 - Strong:** Mostly appropriate and coherent; minor gaps in clarity, depth, or completeness.
   - **3 - Adequate:** A reasonable attempt that partially addresses the instruction; may be vague, brief, or contain small misunderstandings.
   - **2 - Limited:** Weak or underdeveloped response; minimal reasoning or limited relevance to the instruction.
   - **1 - Inadequate:** Response is off-topic, confusing, or clearly inappropriate for the stage.

2. **Use stage-specific developmental expectations**:
   - **Stage 0 (Age 5):** Very simple sentences, concrete ideas, focused on here and now
   - **Stages 1-3 (Ages 6-8):** Simple reasoning, some past/future thinking, familiar examples
   - **Stages 4-6 (Ages 9-11):** Logical structure, comparisons, abstract or hypothetical reasoning
   - **Stages 7-9 (Ages 12-14):** Nuanced reasoning, multi-step thinking, advanced vocabulary

3. **Evaluate:**
   - Does the child's response meaningfully address the instruction?
   - Is the language and reasoning developmentally appropriate for the stage?
   - Is the response authentic and logically consistent?

4. **Output Format:**
Only return the following dictionary:
```json
{{
    "rating": <integer from 1 to 5>,
    "explanation": "<2-3 sentence rationale>"
}}
```
Do not add any other text or formatting. Only return the JSON object.
\end{lstlisting}
User prompt for evaluating trained model's response for questions from IR:
\begin{lstlisting}[style=prompt]
Evaluate the child's response to the instruction below based on the developmental stage and age group. Return a numerical rating (1-5) and a short explanation.

Instruction: {instruction}
Response: {response}
Stage: {stage}
Age group: {age_group}
Index: {q_index}

**Output Format:**
Only return the following dictionary:
```json
{{
    "rating": <integer from 1 to 5>,
    "explanation": "<2-3 sentence rationale>"
}}
```
\end{lstlisting}

\end{document}